\documentclass[letterpaper, 10 pt, conference]{ieeeconf}  % Comment this line out if you need a4paper

\IEEEoverridecommandlockouts                              % This command is only needed if 
\usepackage{graphics} % for pdf, bitmapped graphics files
\usepackage{epsfig} % for postscript graphics files
\usepackage{mathptmx} % assumes new font selection scheme installed
\usepackage{times} % assumes new font selection scheme installed
\usepackage{amsmath} % assumes amsmath package installed
\usepackage{amssymb}  % assumes amsmath package installed
\usepackage{booktabs}
\usepackage{multirow}
\usepackage{caption}
\usepackage{subcaption}
\newcommand{\etal}{\textit{et al}.}

\newcommand{\vs}{\textit{v}.\textit{s}.~}
\usepackage{color}
\definecolor{myblue}{rgb}{0,0,0} 
\newcommand{\zl}[1]{\textcolor{myblue}{#1}}
\usepackage{etoolbox}
\makeatletter
\patchcmd{\@makecaption}
  {\\}
  {.\ }
  {}
  {}
\makeatother

\title{\LARGE \bf
Unseen Object Instance Segmentation with Fully Test-time\\ RGB-D Embeddings Adaptation
}

\author{Lu Zhang, Siqi Zhang, Xu Yang, Hong Qiao and Zhiyong Liu$^{*}$% <-this % stops a space
\thanks{All authors are with State Key Laboratory of Multimodal Artificial Intelligence Systems, Institute of Automation, Chinese Academy of Sciences, Beijing, China, and also with the School of Artificial Intelligence, University of Chinese Academy of Sciences, Beijing, China. }%
\thanks{Zhiyong Liu is also with Nanjing Artificial Intelligence Research of IA, Jiangning District, Nanjing, 211100, Jiangsu, China.}%
\thanks{\{\tt\footnotesize lu.zhang@ia.ac.cn, zhiyong.liu@ia.ac.cn\}}%
}

\begin{document}

\maketitle
\thispagestyle{empty}
\pagestyle{empty}

%%%%%%%%%%%%%%%%%%%%%%%%%%%%%%%%%%%%%%%%%%%%%%%%%%%%%%%%%%%%%%%%%%%%%%%%%%%%%%%%
\begin{abstract}
Segmenting unseen objects is a crucial ability for the robot since it may encounter new environments during the operation. Recently, a popular solution is leveraging RGB-D features of large-scale synthetic data and directly applying the model to unseen real-world scenarios. However, the domain shift caused by the sim2real gap is inevitable, posing a crucial challenge to the segmentation model. In this paper, we emphasize the adaptation process across sim2real domains and model it as a learning problem on the BatchNorm parameters of a simulation-trained model. Specifically, we propose a novel non-parametric entropy objective, which formulates the learning objective for the test-time adaptation in an open-world manner. Then, a cross-modality knowledge distillation objective is further designed to encourage the test-time knowledge transfer for feature enhancement. Our approach can be efficiently implemented with only test images, without requiring annotations or revisiting the large-scale synthetic training data. Besides significant time savings, the proposed method consistently improves segmentation results on the overlap and boundary metrics, achieving state-of-the-art performance on unseen object instance segmentation.

\end{abstract}

%%%%%%%%%%%%%%%%%%%%%%%%%%%%%%%%%%%%%%%%%%%%%%%%%%%%%%%%%%%%%%%%%%%%%%%%%%%%%%%%
\vspace{-0.5em}
\section{INTRODUCTION}
\vspace{-0.5em}
In recent years, a rising development trend in robotics is the transition from controlled labs to unstructured environments. When encountering new environments and objects, the robot system must have the ability to adjust itself and recognize unseen objects. Such capability is essential for robots to understand working environments better and perform various manipulation tasks. To achieve this goal, we approach the task of Unseen Object Instance Segmentation (UOIS) \cite{xie2019best, xiang2020learning, xie2021rice, xie2021unseen}, which aims to conduct instance-aware segmentation of unseen objects in tabletop scenes. In UOIS, the robot system needs to learn the concept of ``object'' and generalize it to unseen ones.

However, unlike ImageNet \cite{deng2009imagenet} and MS COCO \cite{lin2014microsoft}, which have spurred significant development of the classification and object detection for natural images, a large-scale realistic dataset that contains sufficient objects for robotic manipulation scenes is currently unavailable \cite{xiang2020learning}. As a result, existing UOIS methods \cite{xie2019best, xiang2020learning, xie2021rice, xie2021unseen} generally resort to large-scale synthetic RGB-D data to train a perception model. For example, Xie \etal ~\cite{xie2019best} propose using synthetic scenes that can be rendered into RGB-D images from various viewpoints with automatically generated labels. Previous works \cite{xie2019best, xiang2020learning, xie2021rice, xie2021unseen} typically use synthetic depth or RGB-D images to train a model that can separately segment each object instance, since the depth inputs have better generalization potential than the non-photorealistic RGB images. After training, the model is directly deployed on unseen realistic datasets. Though such a workaround achieves good performance, it has in turn led to the neglect of domain shift caused by the ``sim2real gap''. For robot perception, the synthetic data fail to model many aspects of the real world, like the object texture, lighting conditions, depth noises, etc. This sim2real gap degrades the model's performance on realistic data, especially on those in unseen environments.

Therefore, rather than betting the generalization ability to elaborate models as in previous works \cite{xie2019best, xiang2020learning, xie2021rice, xie2021unseen}, we place emphasis on the model's adaptability across domains. 
Specifically, we propose a Fully Test-time RGB-D Embeddings Adaptation (FTEA) framework to mitigate the domain gap between synthetic and unseen realistic data. First, we propose a novel Non-parametric Entropy Objective (NEO) that can be calculated without the explicit classification layer to enable test-time adaptation for the open-world UOIS task. The NEO leverages non-parametric distributions to formulate Shannon entropy \cite{shannon1948mathematical} as the learning objective, since the Shannon entropy is shown to be related to error and shift \cite{wang2020tent}, \textit{i.e.,} more confident predictions are generally more correct and have fewer shifts. Second, we design a Cross-modality Knowledge Distillation (CKD) module to encourage knowledge transfer during testing. Different from other KD methods \cite{hinton2015distilling, gou2021knowledge}, CKD aims to distill the knowledge from multimodal (RGB-D) features to unimodal (RGB or depth) ones, thus enhancing the unimodal features for better fusion. Finally, we utilize the affine transformation provided by BatchNorm (BN) \cite{ioffe2015batch} layer as the modulation parameters to conduct fully test-time RGB-D embeddings adaptation. 
    
\zl{Given the simulation-trained model, FTEA is independent of re-training on the large-scale synthetic data} and does not introduce extra parameters, which establishes a key advantage of flexibility. Meanwhile, compared to the inference process (taking 1$\sim$10s per frame in recent state-of-the-arts \cite{xiang2020learning, xie2021rice}), the computation overhead of the proposed adaptation is negligible (taking 0.04s per frame with a total of 500 iterations), making FTEA particularly efficient. The proposed method is evaluated on two real-world RGB-D image datasets for the unseen object instance segmentation, \textit{i.e.,} OSD \cite{richtsfeld2012segmentation} and OCID \cite{suchi2019easylabel}. Extensive experiments show that FTEA consistently improves segmentation performances and achieves state-of-the-art results on various evaluation metrics, demonstrating the effectiveness of our method.

\begin{figure*}
\centering
\includegraphics[width=6.6in]{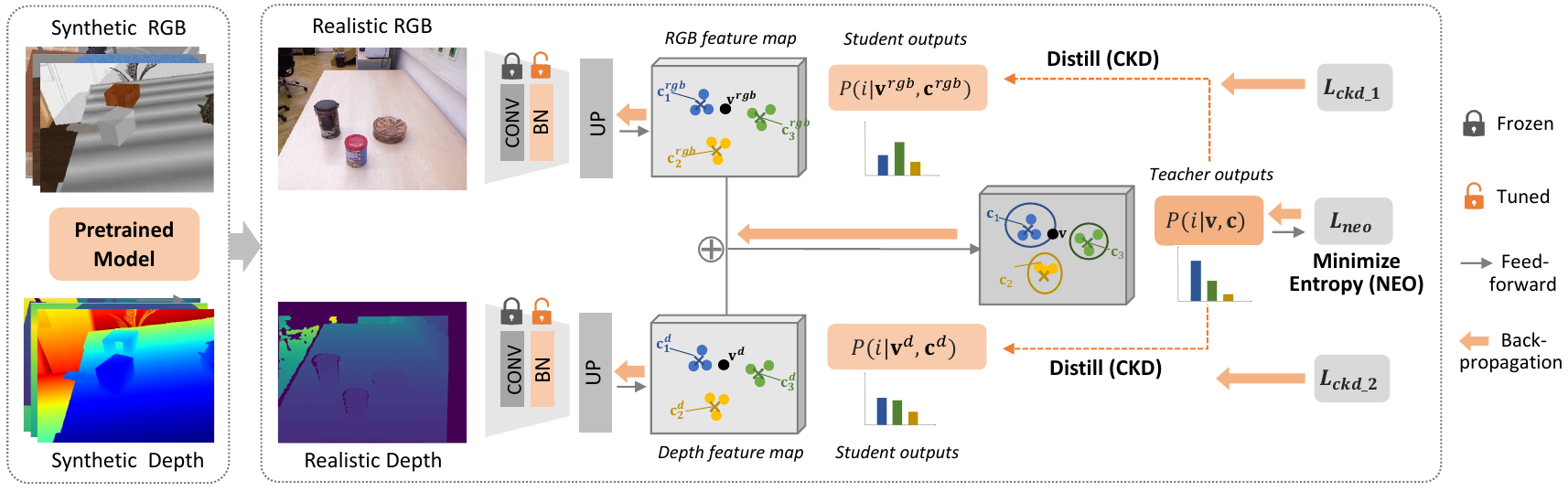}   %
%\captionsetup{font={footnotesize, stretch=1.0}}
\caption{Illustration of the proposed FTEA. We use the two-stream CNN architecture for RGB-D inputs, which is largely simplified for better visualization. During test time, all convolutional layers are frozen, only the affine parameters and normalization statistics of the BN layer are modulated based on two novel objectives, \textit{i.e.,} NEO and CKD. }% 
\label{figure-framework}                     
\end{figure*}

\section{RELATED WORK}
\textbf{Unseen Object Instance Segmentation} UOIS aims to conduct instance-aware segmentation of unseen objects in tabletop environments, which is useful for robots to perform various manipulation tasks. As a pioneer, Xie \textit{et al.}~\cite{xie2019best, xie2021unseen} tackle this problem by proposing a two-stage framework. The framework first operates only on depth to produce rough initial segmentation masks and then refines those masks with RGB. Then, Xiang \textit{et al.}~\cite{xiang2020learning} introduce a fully convolutional network based model called UCN that can be trained end-to-end. Different from previous approaches that mainly rely on depth for segmentation, UCN utilizes both depth images and non-photorealistic RGB images to produce feature embeddings for every pixel, which can be used to learn a distance metric for clustering to segment unseen objects. Recently, several methods are proposed to tackle specific challenges in UOIS. For instance, RICE \cite{xie2021rice} focuses on the occlusion problem in clutter scenes and utilizes a graph-based representation of instance masks to refine the outputs of previous methods. UOAIS-Net \cite{back2021unseen} presents a new unseen object amodal instance segmentation (UOAIS) task to emphasize the amodal perception for robotic manipulation in a cluttered scene, and introduce a large-scale photo-realistic synthetic dataset named UOAIS-SIM to improve the sim2real transferability. \zl{Other works related to UOIS include clustering \cite{tian2017deepcluster, caron2018deep, hsu2015neural}, novel category discovery \cite{hsu2018learning, hsu2018multi, han2019learning, han2021autonovel}, and unsupervised semantic segmentation \cite{ji2019invariant, ouali2020autoregressive, van2021unsupervised}.}

Though \zl{previous UOIS} methods have been demonstrated to be effective, the domain shift problem is not explicitly concerned and tackled. It is worth noting that the photo-realistic data used by UOAIS-Net are also generated with rendered scenes in the simulator, thus synthetic. Differently, we present a new perspective for the solution of the domain shift problem in UOIS and propose an efficient framework to conduct the adaptation during test time.

\textbf{Test-time Adaptation} When the model is deployed, it is inevitable to encounter unlabelled images that are not observed before. This is a key characteristic of robot perception and the UOIS task. Therefore, developing strategies to adapt the model at test time is essential. Recently, test-time training (TTT, TTT+)~\cite{sun2020test, liu2021ttt++} uses unlabelled test instances to conduct self-supervised learning as the adaptation. But such methods heavily rely on the choice of proxy tasks and also need to visit training data that could be unavailable in practice. To address the above limitations, Wang \etal~\cite{wang2020tent} propose Tent to reduce generalization error by test-time entropy minimization. Tent optimizes itself according to its own predictions, which is not relevant to proxy tasks. However, the aforementioned test-time adaptation method is mainly applied in traditional close-set tasks such as image classification, and exploits the output distributions of classifiers as the test-time objective optimization. For the UOIS task, the model generally can not be trained with an explicit discriminative layer with a certain number of classes, which poses a major obstacle for the test-time adaptation in UOIS.
\section{METHOD}
%\vspace{-0.8em}
\subsection{Overview}
%\vspace{-0.3em}
\subsubsection{Network Architecture}
To deal with RGB-D inputs, we adopt the two-stream CNN with late fusion as our basic network architecture. As shown in Figure \ref{figure-framework}, a pair of realistic RGB-D images are separately processed with CNN, and the RGB and depth feature maps are bilinearly upsampled to the full resolution as the input images. Then the late fusion is conducted for a joint representation. We use the UCN \cite{xiang2020learning} as our simulation-trained model due to its conciseness and end-to-end fashion.

\subsubsection{The Pipeline}
First, we construct a non-parametric entropy objective (NEO) for test-time adaptation in an open-world setting, as described in Section \ref{sec3.2}. Then, in Section \ref{sec3.3}, a cross-modality knowledge distillation (CKD) module is further proposed to encourage test-time knowledge transfer for feature enhancement. Finally, we fix all convolutional layers and minimize the proposed NEO and CKD loss to modulate the affine parameters and normalization statistics of the BN layer, as detailed in Section \ref{sec3.4}.

\subsection{Non-parametric Entropy Objective}
\label{sec3.2} 
To modulate features during test time, a learning objective based on the model's predictions of test data is typically required. Recent works \cite{wang2020tent, mummadi2021test} have demonstrated the effectiveness of using the entropy objective based on discriminative outputs (\textit{e.g.}, classification probabilities) \cite{wang2020tent, mummadi2021test}. However, unlike the standard recognition model, UOIS is conducted in an open-world setting, and thus does not explicitly train a discriminative layer that generate logits and probabilities for the direct calculation of entropy. To address this problem, we propose a novel non-parametric entropy objective (NEO). Specifically, NEO leverages the unsupervised clustering to obtain centroids that present pseudo instance labels, then calculates non-parametric classification probabilities to construct the entropy objective. 
    
\textbf{Unsupervised Clustering} Given a bunch of RGB-D embeddings on a feature map, we aim to cluster all pixels into groups to segment unseen objects. But the number of unseen objects is uncertain, which prevents the usage of clustering algorithms with a known number of clusters such as $k$-means or spectral clustering. Thus, we follow UCN \cite{xiang2020learning} to use the mean shift \cite{comaniciu2002mean} clustering algorithm with the von Mises-Fisher (vMF) distribution \cite{kobayashi2010mises}, which automatically discovers the number of objects and generates a segmentation mask for each object. After the unsupervised clustering, we can calculate the centroid of each cluster. The $i$-th cluster's centroid vector $\mathbf{c}_i$ is obtained by averaging all feature map vectors $\mathbf{v}_{x, y}$ which belong to the $i$-th cluster $\mathbf{C}_{i}$ as
\begin{align}
\mathbf{c}_i = \mathrm{avg}(\mathbf{v}_{x,y})~\mathrm{for~all}~\mathbf{v}_{x,y}\in\mathbf{C}_{i}, ~i=1,2,...,n
\end{align}
where $x$ and $y$ denote locations on the feature map along the $x$-axis and $y$-axis. After performing the average operation for each cluster, we obtain the set of all centroids $\mathbf{c}$ as
\begin{align}
\mathbf{c} = \{\mathbf{c}_1, \mathbf{c}_2, ..., \mathbf{c}_n\},
\end{align}
where $n$ is the number of objects, which is estimated by the unsupervised clustering algorithm.

\textbf{Non-parametric Classification Probability} Different from the classification with standard supervised learning, UOIS segments objects in the instance-level, thus having an uncertain number of ``classes'', which prevents the usage of the parametric classifier. Inspired by recent self-supervised learning methods with instance discrimination \cite{wu2018unsupervised, yang2021instance}, we propose to calculate classification probabilities in a non-parametric way. 

Without specifying the number of classes, we conduct non-parametric classification by using the metric between the candidate feature vector $\mathbf{v}$ and the $i$-th cluster's centroid vector $\mathbf{c}_i$ as 
\begin{align}
P(i|\mathbf{v,c})=\frac{\exp(s(\mathbf{v}, \mathbf{c}_i))}{\sum_{i}\exp(s(\mathbf{v}, \mathbf{c}_i))}, ~
s(\mathbf{v}_1, \mathbf{v}_2)=\frac{\mathbf{v}_{1}^\top \mathbf{v}_2}{\Vert \mathbf{v}_{1}\Vert\Vert \mathbf{v}_2\Vert},\zl{~i\in N_{k},}
\label{eq_ncp}
\end{align}
where $s(\cdot, \cdot)$ is the cosine similarity to measure how well $\mathbf{v}$ matches the $i$-th cluster/object, \zl{$N_{k}$ is the set of numbers to indicate candidate $\mathbf{v}$'s corresponding cluster and $k$ nearest clusters.}
    
\textbf{Entropy Objective} As an unsupervised objective, the Shannon entropy \cite{shannon1948mathematical} is widely used and demonstrated to be effective for assessing the output error and shift \cite{wang2020tent}. With the proposed non-parametric classification probability, we can calculate the Shannon entropy as 
\begin{align}
L_{neo} = -\sum_{i}P(i|\mathbf{v,c})\log_2 P(i|\mathbf{v,c}),
\label{eq_neo}
\end{align}
where $P(i|\mathbf{v,c})$ is the probability that $\mathbf{v}$ is recognized as the $i$-th cluster given a collection of centroids $\mathbf{c}$.

\subsection{Cross-modality Knowledge Distillation}
\label{sec3.3}
Compared to the ``perfectly'' generated RGB-D data, real-world RGB-D images have inevitable noise, such as the black holes in the depth images. The fusion of RGB-D features generally makes the network more stable to such noise \cite{xiang2020learning, huang2021makes}. Therefore, while testing on realistic RGB-D images, the \textit{privileged} \cite{gao2019privileged} multimodal (\textit{i.e.}, RGB-D fused) features can be utilized to enhance the unimodal (\textit{i.e.}, RGB or depth) networks. 
Based on this motivation, we propose to distill the knowledge from the multimodal features to the unimodal ones to encourage the test-time knowledge transfer for feature enhancement. This cross-modality knowledge distillation (CKD) constructs another objective for test-time adaptation. In CKD, the full multimodal network (two-stream CNN) is set as the teacher, while the student is the smaller partial unimodal network (one-stream CNN).

We use soft targets (\textit{i.e.,} probabilities of the input belonging to the classes) as the distilled knowledge in CKD. For each candidate feature vector $\mathbf{v}$ on the teacher feature map, the $i$-th soft target is
%\vspace{-0.5em}
\begin{align}
P(i|\mathbf{v,c,}~T)=\frac{\exp(s(\mathbf{v}, \mathbf{c}_i)/T)}{\sum_{i}\exp(s(\mathbf{v}, \mathbf{c}_i)/T)},
\label{eq_distill}
\end{align}
where $T$ is the temperature factor to control the importance of each soft target. Due to the absence of a parametric classifier, here we use the non-parametric probability as similar with Equation \ref{eq_ncp}. 

For the individual RGB and depth modality (\textit{i.e.,} the student), we use the same spatial cluster assignment as the teacher, but they hold different representations $\mathbf{v}^{rgb}$ and $\mathbf{v}^{d}$, thus producing different cluster centroids $\mathbf{c}^{rgb}_i$ and $\mathbf{c}^{d}_i$. Then, the soft targets of RGB and depth modality, \textit{i.e.,} $P(i|\mathbf{v}^{rgb},\mathbf{c}^{rgb},~T)$ and $P(i|\mathbf{v}^{d},\mathbf{c}^{d},~T)$, can be obtained similarly to Equation \ref{eq_distill}. Finally, we formulate the cross-modality knowledge distillation objective as
%\vspace{-0.5em}
\begin{align}
\nonumber L_{ckd}=&\frac{1}{2}(KL(P(i|\mathbf{v,c,}~T), P(i|\mathbf{v}^{rgb},\mathbf{c}^{rgb},~T)) +\\
&KL(P(i|\mathbf{v,c,}~T), P(i|\mathbf{v}^{d},\mathbf{c}^{d},~T))),
\label{eq_ckd}
\end{align}
where $KL(\cdot, \cdot)$ denotes the Kullback-Leibler divergence loss. \zl{The gradients propagated through the multimodal distribution are stopped.}
By optimizing Equation \ref{eq_ckd}, the response knowledge of the multimodal teacher network is distilled into the unimodal student network.

\begin{table*}[!htbp]
\begin{center}
\linespread{1.10}\selectfont
\scalebox{1.1}{
\begin{tabular}{ccc|ccc|ccc|c||ccc|ccc|c}
\toprule  %添加表格头部粗线
\multicolumn{3}{c|}{\multirow{3}*{Method}} &\multicolumn{7}{c||}{OSD \cite{richtsfeld2012segmentation}} &\multicolumn{7}{c}{OCID \cite{suchi2019easylabel}}\\
\cline{4-17}
&&&\multicolumn{3}{c|}{Overlap} & \multicolumn{3}{c|}{Boundary} & ~ & \multicolumn{3}{c|}{Overlap}& \multicolumn{3}{c|}{Boundary}\\
%\cline{5-13}
&&&P &R &F &P &R &F &F@.75 &P &R &F &P &R &F &F@.75\\

\hline

\multicolumn{3}{c|}{Mask RCNN \cite{he2017mask}}&74.4 &72.7 &73.4 &53.1 &48.1 &49.8 &- &80.8 &73.9 &76.1 &68.2 &58.4 &61.8 &- \\

\multicolumn{3}{c|}{UOIS-Net-2D \cite{xie2019best}} &80.7 &80.5 &79.9 &66.0 &67.1 &65.6 &71.9 &88.3 &78.9 &81.7 &82.0 &65.9 &71.4 &69.1\\

\multicolumn{3}{c|}{UOIS-Net-3D \cite{xie2021unseen}} &85.7 &82.5 &83.3 &75.7 &68.9 &71.2 &73.8 &86.5 &86.6 &86.4 &80.0 &73.4 &76.2 &77.2\\

\multicolumn{3}{c|}{UCN \cite{xiang2020learning}} &84.3 &88.3 &86.2 &67.5 &67.5 &67.1 &79.3 &86.0 &92.3 &88.5 &80.4 &78.3 &78.8 &82.2\\

\multicolumn{3}{c|}{UCN+ \cite{xiang2020learning}} &87.4 &87.4 &87.4 &69.1 &70.8 &69.4 &83.2 &91.6 &92.5 &91.6 &\textbf{86.5} &87.1 &86.1 &89.3\\

\multicolumn{3}{c|}{UOAIS-Net \cite{back2021unseen}} &85.3 &85.4 &85.2 &\textbf{72.7} &74.3 &73.1 &79.1 &70.7 &86.7 &71.9 &68.2 &78.5 &68.8 &78.7\\

\hline
\multicolumn{3}{c|}{\textbf{FTEA (Ours)}} &85.8 &\textbf{92.0} &88.6 &69.2 &75.7 &71.7 &87.3 &86.2 &\textbf{93.9} &89.5 &79.5 &79.5 &79.1 &85.1\\
\multicolumn{3}{c|}{\textbf{FTEA+ (Ours)}}&\textbf{89.9} &89.4 &\textbf{89.5} &72.6 &\textbf{76.0} &\textbf{73.8} &\textbf{88.3} &\textbf{92.0} &93.3 &\textbf{92.3} &\textbf{86.5} &\textbf{88.0} &\textbf{86.7} &\textbf{91.1} \\
\bottomrule

\end{tabular}}
\end{center}
%\captionsetup{font={footnotesize, stretch=1.0}}
\vspace{-1em}
\caption{The unseen object instance segmentation (UOIS) performances of the proposed FTEA and other state-of-the-art methods on OSD \cite{richtsfeld2012segmentation} and OCID \cite{suchi2019easylabel} datasets. ``+'' denotes the zoom-in operation \cite{xiang2020learning} to refine segmentation results.}
\label{table-comparisons}
\end{table*}

%\vspace{-0.5em}
\subsection{Fully Test-time RGB-D Embeddings Adaptation}
\label{sec3.4}
%\vspace{-0.5em}
By minimizing the above entropy objective $L_{neo}$ and distillation objective $L_{ckd}$, we can adapt our model fully in test time. However, tuning all parameters like that in the training phase is inefficient and could easily cause model instability \cite{wang2020tent}. As the channel of the feature map can be considered as a feature detector \cite{zeiler2014visualizing, woo2018cbam}, re-calibrating channel responses has been widely studied and utilized in network pruning \cite{he2017channel}, multimodal fusion \cite{wang2020deep, zhang2019cross}, representation learning \cite{hu2018squeeze, shao2020channel}, etc. \zl{Besides, previous works \cite{li2018adaptive, chang2019domain, zhuang2020rethinking, klingner2022unsupervised} on unsupervised domain adaptation (UDA) share a similar idea, \textit{i.e.,} utilizing statistics of the BN layer as domain-related knowledge. However, the UDA methods require the use of source or target data to conduct the adaptation in advance, thus becoming difficult to be applied during test time or online. } \zl{Differently, we use the channel-wise affine transformation provided by the BatchNorm (BN) \cite{ioffe2015batch} layer to stabilize the test-time adaptation and aim for an effective fusion of RGB-D embeddings.} 

In genral, the BN layer is widely used in deep learning to eliminate internal covariate shift and improve generalization. It performs a linear transformation followed by the convolutional or fully-connected layers. We denote by $\mathbf{x}_{m,l,k}$ the $k$-th channel for the $l$-th layer feature maps of $m$-th modality (RGB or depth in our setting), then the transformation of the BN layer can be written as
\vspace{-0.5em}
\begin{align}
\mathbf{x}^{\prime}_{m,l,k}=\gamma_{m,l,k}\frac{\mathbf{x}_{m,l,k}-\mu_{m,l,k}}{\sqrt{\sigma^{2}_{m,l,k}+\epsilon}}+\beta_{m,l,k},
\end{align}
where the scaling and shift factors $\gamma_{m,l,k}$ and $\beta_{m,l,k}$ are adjustable affine parameters, the normalization statistics $\mu_{m,l,k}$ and $\sigma_{m,l,k}$ are updated with momentum.

Thus, by adapting the scaling and shift factors $\gamma_{m,l,k}$, $\beta_{m,l,k}$ and updating the statistics $\mu_{m,l,k}$, $\sigma_{m,l,k}$ of the BN layer, the proposed method does not introduce new parameters. Given the simulation-trained model, our method is independent of re-training on the large-scale synthetic data. Meanwhile, the fully test-time adaptation re-calibrates channel-wise responses of RGB and depth feature maps, thus providing better weighted fused RGB-D embeddings for unseen object instance segmentation. Finally, the overall objective for test-time adaptation can be formulated as
\vspace{-0.5em}
\begin{align}
L_{total} = \lambda_1L_{neo} + \lambda_2L_{ckd},
\label{eqn_loss}
\end{align}
where the two terms $L_{neo}$ and $L_{ckd}$ are weighted by the balancing parameters $\lambda_1$ and $\lambda_2$.

\section{EXPERIMENTS}
\subsection{Datasets and Evaluation Metrics}
\textbf{Datasets} The model is pre-trained on the synthetic Tabletop Object Dataset (TOD) \cite{xie2019best}, which consists of 40k synthetic scenes of cluttered objects in tabletop environments. The proposed method is evaluated and compared on two real-world \zl{datasets}, named OSD \cite{richtsfeld2012segmentation} and OCID \cite{suchi2019easylabel}. OSD consists of 111 images in tabletop environments with averaged 3.3 objects per image. OCID has 2,346  images on both tabletop and floor with averaged 7.5 objects per image. It is worth noting that OSD has manually labeled segmentation masks while OCID generates semi-automatically labeled annotations, which are easily influenced by the noise of depth images.

\textbf{Evaluation Metrics} We follow previous works \cite{xie2019best, xiang2020learning, xie2021unseen} to use the object precision/recall/F-measure (Overlap P/R/F) metrics to evaluate the object segmentation performance, and the Boundary P/R/F to evaluate how sharp the predicted boundary matches against the ground truth boundary. Besides, F@.75 is used to denote the percentage of segmented objects with Overlap F-measure $\geq 75\%$ \cite{xiang2020learning}. Though the IoU is a standard metric in segmentation tasks, it is highly correlated to the overlap F-measure, thus is not reported in the UOIS task \cite{xie2021rice}. All P/R/F and F@.75 measures are reported in the range of $[0, 100]$.

\begin{table*}[!htbp]
\begin{center}
\linespread{1.10}\selectfont
\scalebox{1.1}{
\begin{tabular}{ccc|ccc|ccc|c||ccc|ccc|c}
\toprule  %添加表格头部粗线
\multicolumn{2}{c}{\multirow{3}*{$L_{neo}$}} &\multirow{3}*{$L_{ckd}$} &\multicolumn{7}{c||}{OSD \cite{richtsfeld2012segmentation}} &\multicolumn{7}{c}{OCID \cite{suchi2019easylabel}}\\
\cline{4-17}
&&&\multicolumn{3}{c|}{Overlap} & \multicolumn{3}{c|}{Boundary} & ~ & \multicolumn{3}{c|}{Overlap}& \multicolumn{3}{c|}{Boundary}\\
%\cline{5-13}
&&&P &R &F &P &R &F &F@.75 &P &R &F &P &R &F &F@.75\\

\hline
\multicolumn{2}{c}{~}&~ &84.3 &88.3 &86.2 &67.5 &67.5 &67.1 &79.3 &86.0 &92.3 &88.5 &80.4 &78.3 &78.8 &82.2\\

\multicolumn{2}{c}{$\checkmark$}& &85.0 &91.9 &88.2 &67.6 &74.6 &70.4 &89.8 &85.7 &\textbf{94.0} &89.3 &78.0 &78.9 &78.0 &84.9 \\
\multicolumn{2}{c}{~}&$\checkmark$ &85.3 &89.2 &87.1 &\textbf{70.1} &70.2 &69.8 &81.9 &\textbf{86.2} &92.8 &88.9 &\textbf{80.5} &78.9 &\textbf{79.2} &83.1\\
\multicolumn{2}{c}{$\checkmark$}&$\checkmark$&\textbf{85.8} &\textbf{92.0} &\textbf{88.6} &69.2 &\textbf{75.7} &\textbf{71.7} &\textbf{87.3} &\textbf{86.2} &93.9 &\textbf{89.5} &79.5 &\textbf{79.5} &79.1 &\textbf{85.1} \\
\bottomrule

\end{tabular}}
\end{center}
\vspace{-1.0em}
%\captionsetup{font={footnotesize, stretch=1.0}}
\caption{Ablation studies of the proposed FTEA. The ablation of test-time adaptation (TTA) is implicitly included in the first row since it should be in conjunction with at least one objective function, \textit{i.e.,} $L_{neo}$ or $L_{ckd}$.}
\label{table-ablations}
\end{table*}

\begin{figure*}[htbp]
\centering
\begin{minipage}[t]{0.68\textwidth}
\centering
\includegraphics[width=11.5cm]{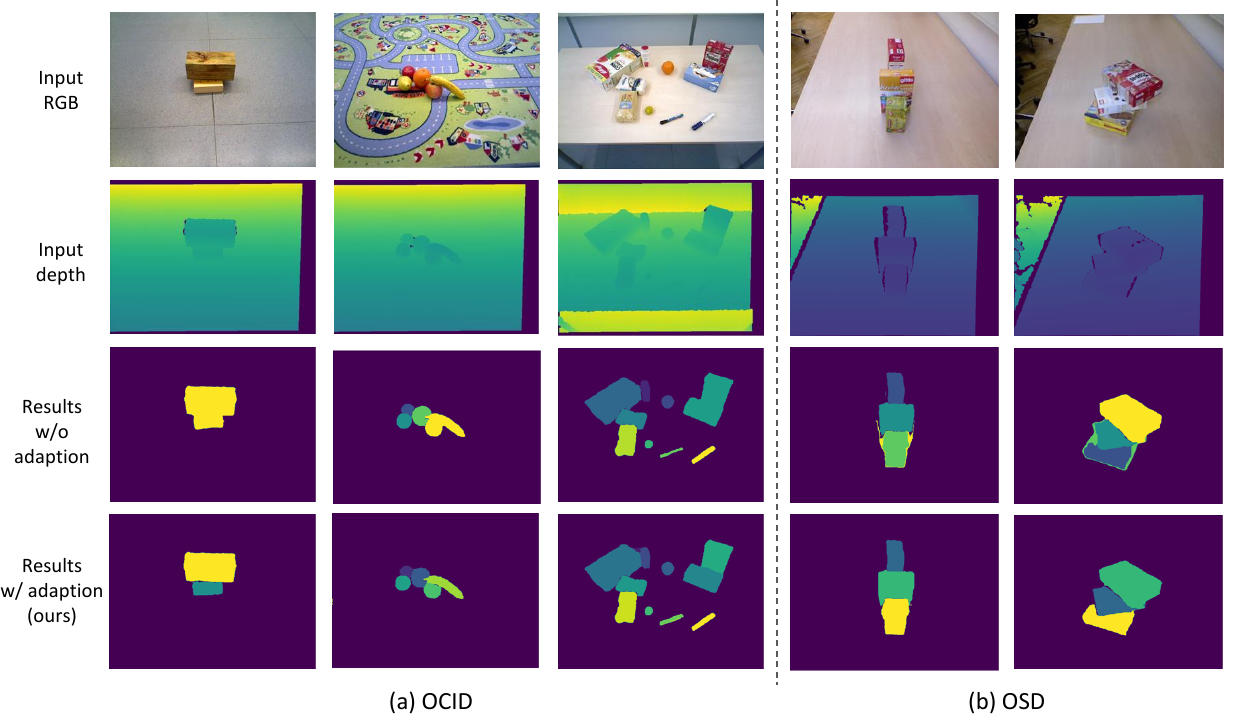}
\vspace{-0.5em}
%\captionsetup{font={footnotesize, stretch=1.0}}
\caption{Qualitative results of the proposed approach, (a) results on the OCID dataset, and (b) results on the OSD dataset.}
\label{figure-visualization}   
\end{minipage}
\begin{minipage}[t]{0.25\textwidth}
\centering
\includegraphics[width=4.0cm]{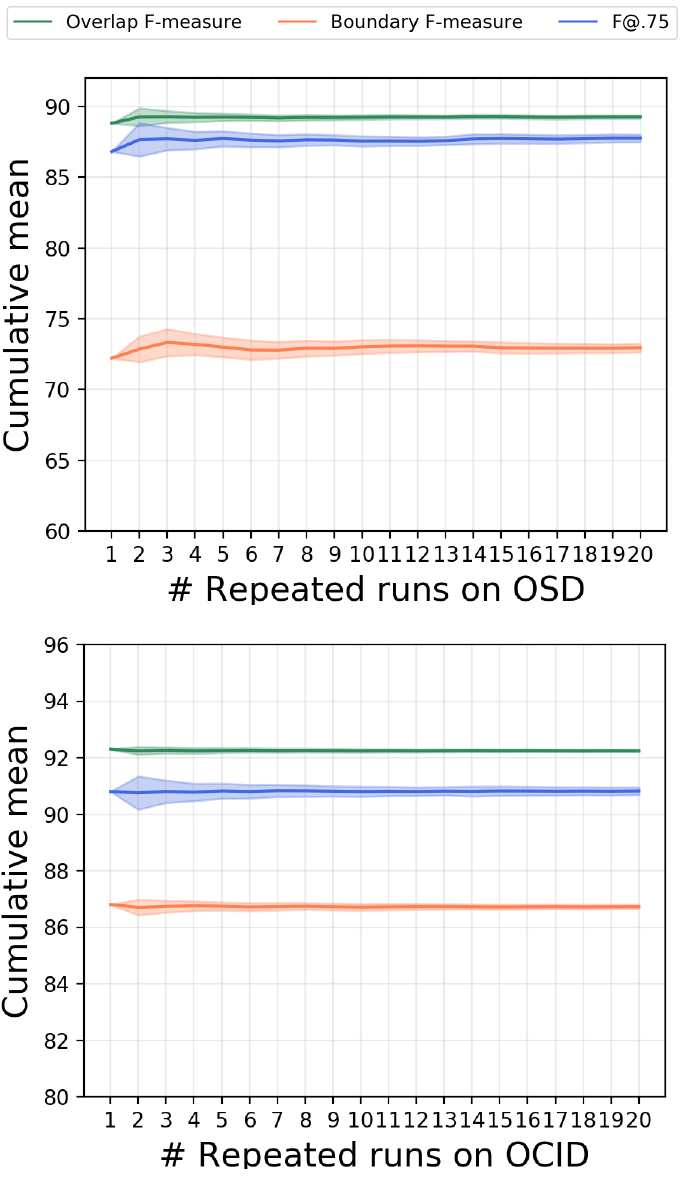}
\vspace{-0.5em}
%\captionsetup{font={footnotesize, stretch=1.0}}
\caption{20 Repeated runs of different sampling orders.}
\label{figure-repeated}
\end{minipage}
\end{figure*}

\subsection{Comparison with State-of-the-art Methods}
In this section, we first compare the proposed method with several state-of-the-art methods. As shown in Table \ref{table-comparisons}, our method outperforms all competitors on both OSD and OCID datasets. Specifically, FTEA+\footnote{FTEA+ denotes adopting the zoom-in refinement strategy in \cite{xiang2020learning}.} achieves 89.5 and 92.3 Overlap F-measure, 73.8 and 86.7 Boundary F-measure, 88.3 and 91.1 F@.75, on OSD and OCID respectively, which validates the effectiveness of the proposed approach. Additionally, when we use the end-to-end model without extra zoom-in refinement, \textit{i.e.,} FTEA in the next-to-last line in Table \ref{table-comparisons}, the improvement of our method is consistent. Compared to UCN \cite{xiang2020learning}, the proposed FTEA improves the Overlap F-measure by 2.4 and 1.0, the Boundary F-measure by 4.6 and 0.3, the F@.75 by 5.1 and 1.8, on OSD and OCID respectively.

\vspace{-0.7em}
\subsection{Ablation Studies}
\vspace{-0.3em}
\textbf{Non-parametric Entropy Objective} This objective is crucial for UOIS to conduct test-time adaptation. As shown in Table \ref{table-ablations}, when equipped with the non-parametric entropy objective $L_{neo}$, the model's performances on two real-world datasets are overall improved. Especially, $L_{neo}$ significantly improves the overlap and boundary \textit{recall}, which is desirable for discovering potential objects in unseen environments.

\textbf{Cross-modality Knowledge Distillation} The proposed cross-modality knowledge distillation provides another effective learning objective $L_{ckd}$ for test-time feature enhancement. Table \ref{table-ablations} shows that $L_{ckd}$ works well on improving the overlap and boundary \textit{precision} of segmentation results. Additionally, as shown in the last row in Table \ref{table-ablations}, the cross-modality knowledge distillation $L_{ckd}$ can be effectively combined with the former entropy objective $L_{neo}$, thus further improving the performance on most evaluation metrics for unseen object instance segmentation.

\textbf{Test-time Adaptation} Since the test-time adaptation (TTA) needs to be enabled with at least one proposed learning objective, \textit{i.e.,} NEO or CKD, the ablation of TTA is implicitly included in the first row in Table \ref{table-ablations}. From Table \ref{table-ablations} we can observe that whether TTA is in conjunction with the $L_{neo}$ or $L_{ckd}$, the segmentation results on both OSD and OCID can be significantly improved.

\vspace{-0.5em}
\subsection{Discussions}
\vspace{-0.3em}

\textbf{Adaptation Consumption} Besides the performance, we further investigate the computation consumption of the proposed method. The adaptation time only accounts for single backward pass of BN parameters, thus is way faster than the inference with multiple zoom-in operations \cite{xiang2020learning}. Table \ref{table-consumptations} shows the averaged time consumption of the adaptation process and inference, \textit{i.e.,} averaged 0.04s \vs 1.24s per iteration. Based on the setting of only adapting 500 iterations as stated in Section \ref{sec4.2}, the total adaptation consumption for a new scenario is $\sim$20s (0.04s per frame), which is efficient for practical application. 

%%% TABLE-consumptions
\begin{table}[h]
\begin{center}
\linespread{1.0}\selectfont
\scalebox{1.0}{
\begin{tabular}{c|c|c}
\toprule  %添加表格头部粗线
BatchSize &Inference time (avg) &Adaptation time (avg) \\

\hline
1&1.24$\pm$0.03s&0.04$\pm$0.00s\\

\bottomrule
\end{tabular}}
\end{center}
\vspace{-1em}
%\captionsetup{font={footnotesize, stretch=1.0}}
\caption{The time consumption of the inference and adaptation process, which is averaged over 500 iterations with 10 repeated runs.}
\label{table-consumptations}
\end{table}

\textbf{Qualitative Results} We visualize some qualitative results with and without the proposed FTEA in Figure \ref{figure-visualization}. We can observe in Figure \ref{figure-visualization}(a) that the proposed adaptation process mitigates the under-segmentation problem of two close objects. Besides, different from the smooth depth images in synthetic training data, realistic depth images are generally noisy, especially on the object's boundary. This problem makes outputs of boundaries blurred and causes over-segmentation around the boundary, as illustrated in Figure \ref{figure-visualization}(b). The last row in Figure \ref{figure-visualization}(b) shows that, with the proposed adaptation process in FTEA, this problem can be largely alleviated.

%%% TABLE-modalities
\begin{table}[t]
\begin{center}
\linespread{1.05}\selectfont
\scalebox{1.1}{
\begin{tabular}{cc|c|c|c}
\toprule  %添加表格头部粗线
\multicolumn{2}{c|}{Modalities} &Overlap &Boundary& \\
RGB & Depth &F &F &F@.75  \\
\hline
& &87.4 &69.4 &83.2\\
$\checkmark$& &87.6 &70.0 &86.1\\
&$\checkmark$ &88.6 &70.9 &87.4\\
$\checkmark$ &$\checkmark$ &\textbf{89.5} &\textbf{73.8} &\textbf{88.3}\\
\midrule
\multicolumn{2}{c|}{\zl{Parameters}} &\zl{Overlap} &\zl{Boundary}& \\
\zl{Conv} & \zl{BN} &\zl{F} &\zl{F} &\zl{F@.75}  \\
\hline
& &\zl{87.4} &\zl{69.4} &\zl{83.2}\\
\zl{$\checkmark$}& &\zl{87.4} &\zl{69.4} &\zl{83.2}\\
&\zl{$\checkmark$} &\zl{\textbf{89.5}} &\zl{\textbf{73.8}} &\zl{\textbf{88.3}}\\
\zl{$\checkmark$} &\zl{$\checkmark$} &\zl{82.2} &\zl{65.4} &\zl{75.6}\\
\bottomrule
\end{tabular}}
\end{center}
\vspace{-1.0em}
%\captionsetup{font={footnotesize, stretch=1.0}}
\caption{Segmentation performances with different adaptation modalities \zl{and modulation parameters} on OSD.}
\label{table-modalities}
\end{table}

\textbf{Robustness to the Sampling Order}
Additionally, we train our models for multiple runs on different random sampling orders. Figure \ref{figure-repeated} illustrates the cumulative means of performance with $95\%$ confidence intervals across 20 repeated runs on OCID and OSD. It can be observed that the overall performances of the model are stable, demonstrating the robustness of the proposed method. Note that the experiments on OCID also use random samples (not just the sampling order) since we conduct the adaptation with only 500 test-time images out of 2,346 images in OCID.

\textbf{Modalities for Adaptation} Table \ref{table-modalities} shows results with different adaptation modalities on OSD. First, segmentation performances can be improved whether RGB or depth modality is tuned.
Second, by simultaneously tuning RGB and depth modalities, we can obtain better-fused RGB-D embeddings. Thus the performances become significantly better (see the last row in Table \ref{table-modalities}) due to the effective use of multimodal data.

\zl{\textbf{Modulation Parameters}} \zl{We also conduct analysis on the modulation parameters, \textit{i.e.}, the linear BN and the nonlinear convolution parameters. As shown in Table \ref{table-modalities}, using BN as modulation parameters achieves the best results. Since the nonlinear high dimensional convolution layers are difficult to optimize, tuning all model parameters could lead to a negative transfer.}

\begin{figure}[tbp]
\centering
\includegraphics[width=2.2in]{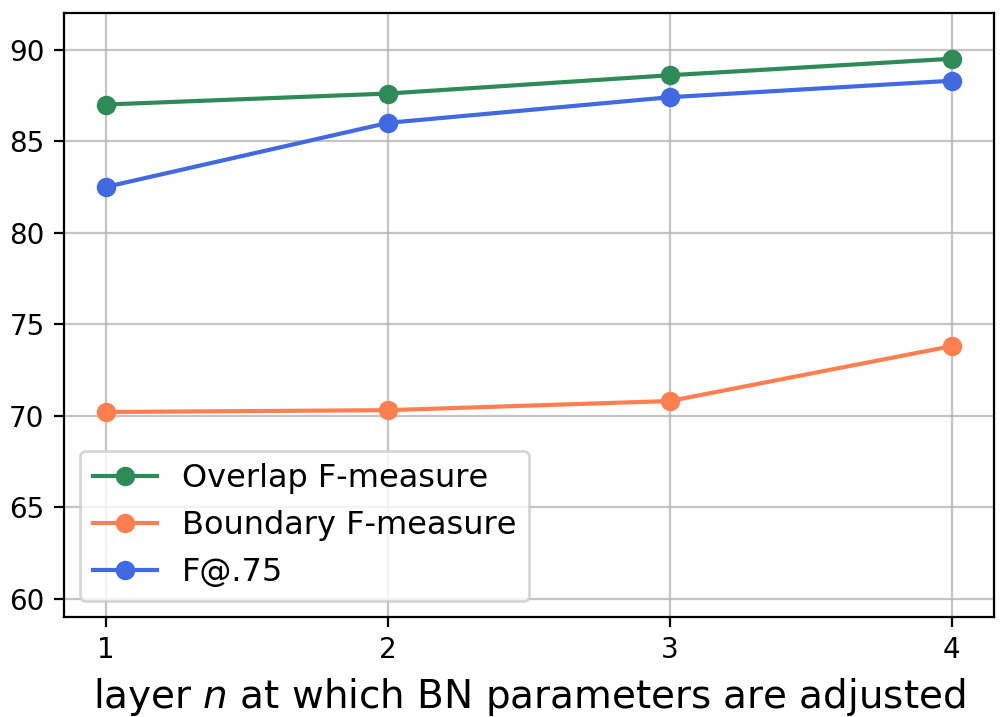}   %
%\captionsetup{font={footnotesize, stretch=1.0}}
\vspace{-0.5em}
\caption{Performances evolution on OSD when we gradually increase the number of BN layers for adaptation.}% 
\label{figure-num-bn}                     
\end{figure}

\textbf{How Many BN Layers to Use}
For brevity, we split all layers into four main building blocks as in ResNet \cite{he2016deep}. Figure \ref{figure-num-bn} illustrates the segmentation performances on OSD when we gradually increase the number of BN layers for adaptation. We can observe that the Overlap F-measure, Boundary F-measure, and F@.75 are consistently improved with more BN layers, demonstrating the effectiveness of the proposed method. On the other hand, the improvement is not plateaued, which implies that the adaptation process can be further enhanced by considering other effective parameters in addition to the BN layers of the model.

\vspace{-0.35em}
\subsection{Implementation Details}
\vspace{-0.15em}
\label{sec4.2}
For fair comparisons, we follow previous work \cite{xiang2020learning} to use a 34-layer, stride-8 ResNet (ResNet34-8s) as the backbone, and the full resolution $640\times480$ feature map with embedding dimensions $C=64$ is obtained by bilinearly upsampling. To avoid noisy outliers, we set $k=1$ in \zl{Equation \ref{eq_ncp}} (\textit{i.e.,} \zl{select two clusters, the nearest cluster and the corresponding cluster)} for the calculation of the non-parametric entropy objective $L_{neo}$. 
The temperature factor $T$ in cross-modality knowledge distillation is 1. The weight factor for the overall loss is set as $\lambda_1=\lambda_2=1$. %The momentum of BN layers is set to 0.5. 
During test time, our model is adapted with the SGD optimizer. We use batchsize=1 as in the typical inference phase. For the first 100 iterations \zl{(images)}, the learning rate is linearly warmed up to the base value $lr=0.005$, then decayed with a cosine scheduler for another 400 iterations \zl{(images)}. \zl{We do not set the ``epoch'' number since there is no concept of ``dataset'' in the online test-time adaptation. We use the same learning schedule for the OSD and OCID. The data in test-time adaptation are not shuffled unless otherwise stated.} All experiments are conducted on a single NVIDIA 2080Ti GPU with PyTorch.

\section{CONCLUSIONS}
In this paper, we target the task of unseen object instance segmentation with an emphasis on the adaptation process for unseen realistic data. To mitigate the domain shift between the synthetic training and realistic testing data, a novel FTEA framework is proposed to conduct the fully test-time RGB-D embeddings adaptation. Specifically, during test time, we fix all convolutional layers and adjust the affine transformations provided by BN parameters via optimizing two novel unsupervised objectives, \textit{i.e.,} the NEO and the CKD. NEO calculates the entropy of probability distributions of UOIS in a non-parametric way. CKD further encourages cross-modality knowledge transfer during test time. Extensive experiments on realistic RGB-D datasets OCID and OSD demonstrate the effectiveness of the proposed approach. We hope our work could draw attention to the test-time adaptation and reveal a promising direction for robot perception in unseen environments.

\section{ACKNOWLEDGEMENT}
 We thank all anonymous reviewers for their constructive suggestions for improving our paper. This work was supported by the NSFC under Grant 62206288; in part by the National Key Research and Development Plan of China under Grant 2020AAA0108902; in part by the Strategic Priority Research Program of Chinese Academy of Science under Grant XDB32050100; in part by the Beijing Science and Technology Plan Project under Grant Z201100008320029; and in part by the Fujian Science and Technology Plan under Grant No. 2021T3003.

\bibliographystyle{IEEEtran}
\bibliography{root}

\end{document}